\documentclass[letterpaper]{article} % DO NOT CHANGE THIS
\usepackage{aaai25}  % DO NOT CHANGE THIS
\usepackage{times}  % DO NOT CHANGE THIS
\usepackage{helvet}  % DO NOT CHANGE THIS
\usepackage{courier}  % DO NOT CHANGE THIS
\usepackage[hyphens]{url}  % DO NOT CHANGE THIS
\usepackage{graphicx} % DO NOT CHANGE THIS
\usepackage{natbib}  % DO NOT CHANGE THIS AND DO NOT ADD ANY OPTIONS TO IT
\usepackage{caption} % DO NOT CHANGE THIS AND DO NOT ADD ANY OPTIONS TO IT
\usepackage{multirow}
\usepackage{bbding}
\usepackage{pifont}
\usepackage{amssymb}
\usepackage{enumitem}
\usepackage[capitalize]{cleveref}
\usepackage{algorithm}
\usepackage{algorithmic}

\usepackage{color}

\usepackage{newfloat}
\usepackage{listings}
\DeclareCaptionStyle{ruled}{labelfont=normalfont,labelsep=colon,strut=off} % DO NOT CHANGE THIS
\floatstyle{ruled}
\newfloat{listing}{tb}{lst}{}
\floatname{listing}{Listing}
\title{Arbitrary Reading Order Scene Text Spotter with Local Semantics Guidance}
\author{
    Jiahao Lyu\equalcontrib\textsuperscript{\rm 1,\rm 4},
    Wei Wang\equalcontrib\textsuperscript{\rm 3},
    Dongbao Yang\textsuperscript{\rm 1},
    Jinwen Zhong\textsuperscript{\rm 1}\thanks{Corresponding Authors.},
    Yu Zhou\textsuperscript{\rm 2}$^\dagger$
}
\affiliations{
    \textsuperscript{\rm 1}Institute of Information Engineering, Chinese Academy of Science\\
    \textsuperscript{\rm 2}VCIP \& TMCC \& DISSec, College of Computer Science, Nankai University\\
    \textsuperscript{\rm 3} Shanghai Artificial Intelligence Laboratory\\
    \textsuperscript{\rm 4}School of Cyber Security, University of Chinese Academy of Sciences\\
}

\usepackage{bibentry}
\begin{document}

\maketitle

\begin{abstract}
Scene text spotting has attracted the enthusiasm of relative researchers in recent years. Most existing scene text spotters follow the detection-then-recognition paradigm, where the vanilla detection module hardly determines the reading order and leads to failure recognition.
After rethinking the auto-regressive scene text recognition method, we find that a well-trained recognizer can implicitly perceive the local semantics of all characters in a complete word or a sentence without a character-level detection module. Local semantic knowledge not only includes text content but also spatial information in the right reading order. Motivated by the above analysis, we propose the Local Semantics Guided scene text Spotter (LSGSpotter), which auto-regressively decodes the position and content of characters guided by the local semantics. Specifically, two effective modules are proposed in LSGSpotter. On the one hand, we design a Start Point Localization Module (SPLM) for locating text start points to determine the right reading order. On the other hand,  a Multi-scale Adaptive Attention Module (MAAM) is proposed to adaptively aggregate text features in a local area. In conclusion, LSGSpotter achieves the arbitrary reading order spotting task without the limitation of sophisticated detection, while alleviating the cost of computational resources with the grid sampling strategy. Extensive experiment results show LSGSpotter achieves state-of-the-art performance on the InverseText benchmark. Moreover, our spotter demonstrates superior performance on English benchmarks for arbitrary-shaped text, achieving improvements of 0.7\% and 2.5\% on Total-Text and SCUT-CTW1500, respectively. These results validate our text spotter is effective for scene texts in arbitrary reading order and shape.
\end{abstract}


\begin{figure}[t]
\centering
\includegraphics[width=0.48\textwidth]{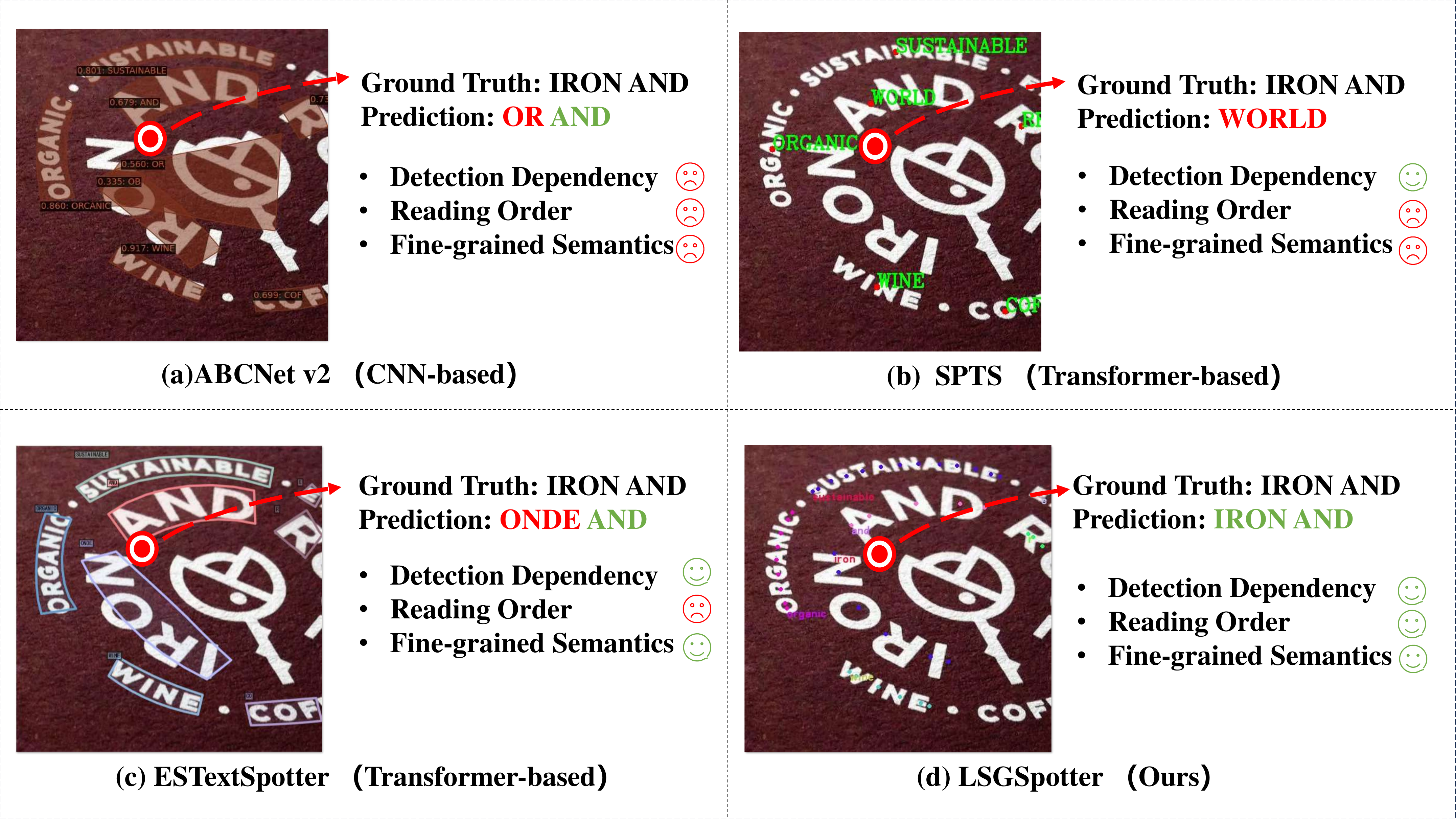}
    \caption{The comparison of arbitrary reading order text instances and analysis from Total-Text   \cite{ch2017total} of ABCNet v2  \cite{liu2021abcnet}, SPTS  \cite{peng2022spts} and ESTextSpotter   \cite{ huang2023estextspotter}. Our spotter can use the locally fine-grained semantics to perceive reading order without accurate detection dependency.}
    \label{fig:problem}
\end{figure}
% \vspace{-20pt}

\section{Introduction}

Aiming to integrate the detection \cite{liao2020real, shu2023perceiving, qin2023towards} and recognition \cite{qiao2020seed, du2022svtr} tasks, scene text spotting has received increasing attention recently because of its numerous applications, such as structure information exaction \cite{ xu2020layoutlm, li2021structext, shen2023divide}, 
automatic driving \cite{guo2021learning, min2022traffic}, scene understanding   \cite{zhu2023locate, zeng2023beyond}, scene text editing \cite{zeng2024textctrl, li2024first} etc. With the development of well-organized datasets \cite{yuliang2017detecting, karatzas2015icdar, ch2017total, ye2023dptext, zhang2019icdar} and fundamental vision models   \cite{dosovitskiy2020image, liu2021swin}, scene text spotters have achieved prominent results on several public benchmarks.

% 总结之前工作 引出检测和识别的协同问题
Existing scene text spotting methods can mainly be divided into two categories according to the utilization of the fundamental vision model: CNN-based and Transformer-based methods. Motivated by general object detection methods  \cite{he2017mask, liu2016ssd}, most of the CNN-based spotters   \cite{lyu2018mask, liao2019mask, liao2020mask, liu2020abcnet, wang2022tpsnet} 
follow the detection-then-recognition paradigm. As the prior stage, detection performance plays a dominant role in the whole pipeline. 
%With advances in the application of Transformers, 
Transformer-based spotters \cite{zhang2022text, huang2022swintextspotter,ye2023deepsolo, huang2023estextspotter} 
%have become the mainstream methods. Transformer-like architectures 
allow the queries of detection and recognition to interact mutually. Some spotters  \cite{peng2022spts, kim2022deer, liu2023spts, kil2023towards} regard the spotting task as sequence generation and try unifying data in multiple OCR-related tasks to improve the performance on scene text spotting.

Although existing spotters achieve remarkable performances, 
%whether CNN-based or Transformer-based spotters, 
arbitrary reading order text spotting is still a challenging problem. As shown in \Cref{fig:problem}, while facing the text instances including the ordinary and inverse texts at the same time, there are obvious spotting failures for all the representative
 methods. ABCNet v2 \cite{liu2021abcnet}, as a representative CNN-based spotter, is powerless to detect inverse text. The detection errors are accumulated and propagated in the recognition stage with Bezier-Align. SPTS \cite{peng2022spts} is a Transformer-based method in an auto-regressive manner. Compared with ABCNet v2, it alleviates the detection dependency and locates the arbitrary reading order instances well. However, SPTS only uses a single point to represent and perceive each text instance due to lacking fine-grained semantics. That is why SPTS produces many incorrect recognition results despite correct localization. As another Transformer-based method, ESTextSpotter \cite{huang2023estextspotter} also reduces the extreme detection dependency. However, it still fails to spot the inverse texts due to the lack of special design.

To solve such failure cases, we revisit how people spot arbitrary reading order texts accurately. People intuitively pay attention to the coarse location of texts and read them character-by-character, similar to how a well-trained auto-regressive recognizer \cite{du2022svtr, qiao2021pimnet} works without accurate character localization \cite{lyu2024textblockv2}.
Motivated by this, we turn the end-to-end text spotting into the recognition problem to alleviate the dependency on detection. 
However, two crucial problems still need to be solved. Firstly, the recognition model hardly identifies the correct reading order. Therefore, a specific module should be designed to determine where to start reading and in what order. Secondly, text instances are scattered in the scene image. If we recognize the features of each character by traversing the whole image, it will waste a lot of computing resources and have low spotting efficiency.

To overcome the problems mentioned above, we propose a Local-Semantics-Guided Scene Text Spotter (LSGSpotter) to handle the arbitrary reading order problem, which exploits the auto-regressive manner elegantly and facilitates the synergy of detection and recognition. Specifically, we design a Start Point Localization Module (SPLM) to separate distinct text instances elaborately. Note that different from the previous detection module, our localization module relaxes the limitation for the recognition stage. The utilization of SPLM also gets the start points and contributes to the correct reading order. 
To solve the second issue mentioned above, a Multi-scale Adaptive Attention Module (MAAM) is proposed to adaptively aggregate text features in a local area. In MAAM, we adopt the strategy of grid sampling to alleviate the computational resource.
During the inference phase, given a scene text image, after extracting the multi-level features using ResNet50 and FPN, the SPLM predicts the start point according to the image feature. As the reference point, the start point guides the feature sampling at the first step during decoding. Then MAAM gets adaptively local feature grids from the reference position and multi-level image features. The cross-attention module of the Transformer decoder can decode the current character and capture the shift of the next character. The above procedure will be repeated until the end-of-sequence token. Therefore, our LSGSpotter can leverage fine-grained semantic information to auto-regressively decode the complete text instance step by step without sophisticated detection dependency. In conclusion, our contributions are as follows:
% 贡献
\begin{itemize}[leftmargin=*]
    \item We propose LSGSpotter, a local semantics-guided scene text spotter to handle the arbitrary reading order text instances without sophisticated detection. Our spotter auto-regressively decodes the position and content of characters guided by the local semantic information to alleviate the dependency on detection.
    \item We introduce two effective modules to solve the arbitrary reading order problem and improve the efficiency of our spotter. Specifically, the Start Point Localization Module (SPLM) aims to localize the reference point for the correct reading order. Guided by local semantics, the Multi-scale Adaptive Attention Module (MAAM) decodes the character shift and content auto-regressively, which enhances the interaction between the position and content information. The strategy of grid sampling in MAAM also releases the computational burden.
    \item Extensive experiments show our proposed method outperforms InverseText, a specific benchmark for arbitrary reading order. Moreover, we also validate the state-of-the-art performances of LSGSpotter on arbitrarily shaped benchmarks, including 81.5\% on Total-Text, and 68.9\% on SCUT-CTW1500 without the help of lexicon.
\end{itemize}

\section{Related Works}

% \begin{figure*}[t]
% \centering
% \includegraphics[width=\textwidth]{ asset/pipeline-refined.pdf}
%     \caption{Overview of several end-to-end scene text spotting methods corresponding to our research objectives. CNN-based spotters and Transformer-based spotters are in blue and yellow areas respectively. Within the ground-truth (GT) box, the letters 'W' and 'C' indicate word-level and character-level annotations, respectively. The figure style follows ABCNet \cite{liu2020abcnet}. Compared with existing methods, ours is more concise in the pipeline.}
%     \label{fig:pipeline}
% \end{figure*}

% Existing scene text spotters can be classified into two categories according to architecture, CNN-based methods and Transformer-based methods. CNN-based methods can be mainly divided into top-down and bottom-up methods. Transformer-based spotters are classified as the non-auto-regressive (NAR) and the auto-regressive (AR) methods according to the difference in the decoding manner.  
% \Cref{fig:pipeline} shows the pipelines of existing scene text spotters and our spotter.

\subsection{CNN-based Methods}
CNN-based spotters are derived from general object detectors, and can mainly be divided into top-down manners and bottom-up manners. 

\subsubsection{Top-down Methods}

For top-down methods, Li et al.  \cite{li2017towards}  firstly propose the end-to-end trainable scene text spotter based on CRNN  \cite{shi2016end}. To solve the arbitrarily shaped text spotting, Mask TextSpotter series  \cite{he2017mask, liao2019mask, liao2020mask} are proposed. Other methods  \cite{lu2022boundary, liu2020abcnet, liu2021abcnet, wang2022tpsnet, qiao2020text} explore various text representations for more accurate text boundaries. AETextspotter \cite{wang2020ae} notices the ambiguity of Chinese layout and introduces the language model to drop out the results of non-semantic character sequences. Top-down spotters adopt the RoI-Align-like methods to align the text features used between detection and recognition prevalently. However, detection-first paradigm methods depend on the accurate detection results extremely. 

\subsubsection{Bottom-up Methods}

Some text spotters try to introduce bottom-up manners to alleviate the detection-dependency problem. CharNet  \cite{xing2019convolutional} and CRAFTS  \cite{baek2020character} use character-level annotations to perform character and text detection in a single pass. MANGO  \cite{qiao2021mango} develops the Mask Attention Module to extract the global features for text instances. PGNet  \cite{wang2021pgnet} performs the text instances with multi-task objectives, such as the centerline, border offset, direction offset and character sequence prediction. Although bottom-up methods eliminate the dependency of detection, they still use a specially designed polygon restoration process and extra character-level annotations.
% consists of the word detection branch and the character segmentation module. The iterative detection scheme transfers the expensive character-level annotations in synthetic datasets into real datasets.

\begin{figure*}[h]
\centering
\includegraphics[width=\textwidth]{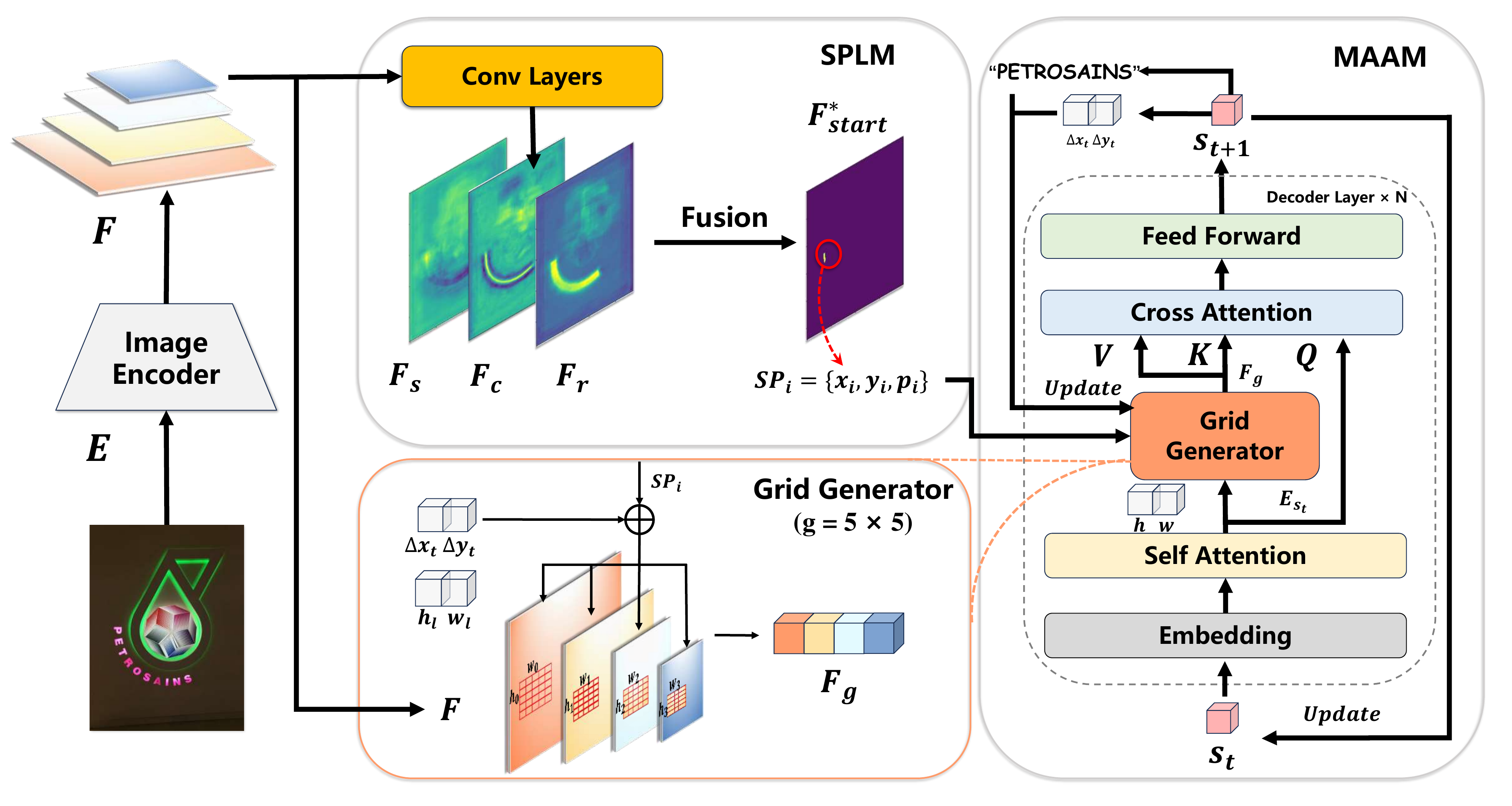}
    \caption{The architecture of LSGSpotter. Image encoder refers to the aggregation of the backbone and neck. SPLM and MAAM are abbreviations of \textit{Start Point Localization Module} and \textit{Multi-level Adaptive Attention Module} respectively. The start point produced by SPLM is the first reference point in the MAAM.}
    \label{fig:model}
    \vspace{-15pt}
\end{figure*}

\subsection{Transformer-based Methods}

\subsubsection{NAR Methods}
With the Transformer's successful applications of various visual tasks, recent works \cite{zhang2022text, huang2022swintextspotter, kittenplon2022towards, huang2023estextspotter} firstly explore the DEtection TRansformer (DETR)  \cite{carion2020end} framework as the main architecture of scene text spotters. Compared with CNN-based methods, Transformer-based methods succeed in long-range modeling and produce a more robust performance on scene text spotting. TTS  \cite{kittenplon2022towards} and TESTR  \cite{zhang2022text} firstly adopt Transformer into the text spotting task. DeepSOLO  \cite{ye2023deepsolo} improves the initialization of queries fed into the SOLO decoder based on TESTR. ESTextSpotter achieves explicit synergy by modeling discriminative and interactive features for text detection and recognition within a single decoder. 
% 总结一下自回归的优劣势

\subsubsection{AR Methods}
AR methods \cite{kim2022deer, peng2022spts, liu2023spts, kil2023towards} model the scene text spotting as a sequence generation task and unify more document tasks. SPTS  \cite{peng2022spts} first proposes to transform the text spotting into the sequence generation task, and later SPTSv2  \cite{liu2023spts} accelerates the inference speed by designing a parallel-decoding scheme. UNITS  \cite{kil2023towards} tries combining with more datasets related to OCR for training a model to balance multiple tasks. Compared with NAR methods, AR methods can organize more data related to OCR tasks for training easily. However, slow inference speed is still an unresolved problem. 

\section{Methodology}

\subsection{Overview}
  \Cref{fig:model} shows the overall architecture of our LSGSpotter. Given a scene text image $I$ with $n$ text instances, an image encoder $E$, including the backbone and neck network, extracts the multi-level feature maps $F$. Next, $F$ is flattened and fed into the Start Point Localization Module (SPLM) to get the start point $SP_i = \{(x_i, y_i, p_i)\}_{i=1}^n$ of each text instance, where $(x_i, y_i)$ and $p_i$ are coordinates of the start point, and the confidence of i-th text instance respectively. After that, the global image features $F$ and start point $SP$ are fed into the Multi-scale Adaptive Attention Module (MAAM) to decode the character content and shift relative to the position of the former character auto-regressively. This section will describe each module of LSGSpotter in detail.

\subsection{Start Points Localization Module}
\label{SPLM}

Generally, scene texts scatter across the whole image. Therefore, to separate different text instances and avoid calculating the global attention of the whole image, we propose a simple Start Points Localization Module (SPLM) to locate the start points.

Given the global multi-level feature $F$, a convolutional block maps $F$ into three channels $F_{s}$, $F_{c}$ and $F_{r}$. The convolutional block consists of 3 Conv-BN-ReLU layers. Specifically, $F_{s}$ is the probability map of text start position, which helps decide the reading order of text instances. $F_{c}$ is the probability map of the text centerline, which assists in separating the adjacent instances. $F_{r}$ is the probability map of the text region. 

\begin{equation}
    F^*_{start} = \sqrt{F_{s} \cdot F_{r} \cdot F_{c}}
    \label{eq:fuse}
\end{equation}

In the inference stage, we fuse three probability maps into a fine-grained start map $F^*_{start}$ as  \Cref{eq:fuse}. Then the start point $SP = \{x_i, y_i, p_i\}_{i=1}^n$ can extracted from $F^*_{start}$, where $(x_i, y_i)$ is the coordinates of the i-th start point and $p_i$ is corresponding confidence. Given a threshold $T$, $n$ connected regions can be generated by $M = \{m_i\}_{i=1}^n = \{\mathbf{1}|F^*_{start} > T\}$. The i-th start point $(x_i, y_i)$ is the center of the i-th connected regions $m_i$, and the confidence $p_i$ is the mean value of probability in $m_i$, termed as $p_i = \{\overline{F(x, y)}|(x, y) \in m_i\}$

During the training stage, polygonal annotations in publicly organized datasets can generate three corresponding ground truths $GT_{s}$, $GT_{c}$ and $GT_{r}$. The detailed label generation method of $GT_{s}$ is described in  \Cref{fig:label}, and $GT_{c}$ and $GT_{r}$ are mentioned by PGNet  \cite{wang2021pgnet}. $F_{r}$ and $F_{c}$ are supervised by BCELoss and $F_{s}$ is supervised by Smooth L1 loss, as shown  \Cref{eq:lr}, (\Cref{eq:lc}) and (\Cref{eq:ls})
. Note that to simplify the calculation of $\mathcal{L}_{s}$ and $\mathcal{L}_{c}$, we only consider the part of the text region. The loss of SPLM is termed as $\mathcal{L}_{start}$, which is the sum of optimization targets $\mathcal{L}_{s}$, $\mathcal{L}_{c}$ and $\mathcal{L}_{r}$ of three maps.

\begin{equation}
    \mathcal{L}_{det} = \mathcal{L}_{s} + \mathcal{L}_{c} + \mathcal{L}_{r},
    \label{eq:start}
\end{equation}

\begin{equation}
    \mathcal{L}_{r} =  \sum_{i=0}^H \sum_{j=0}^{W} BCE(F_{r_{ij}}, GT_{r_{ij}}),
    \label{eq:lr}
\end{equation}

\begin{equation}
\mathcal{L}_{c} = \sum_{i, j \in TR} BCE(F_{c_{ij}}, GT_{c_{ij}}),
    \label{eq:lc}
\end{equation}

\begin{equation}
\mathcal{L}_{s} = \sum_{i, j \in TR} SmoothL1(F_{s_{ij}}, GT_{s_{ij}}).
    \label{eq:ls}
\end{equation}

\subsection{Multi-scale Adaptive Attention Module}
For the scene text recognizer based on Transformer, Self-Attention is responsible for obtaining the semantic information and the cross-attention perceives the visual information of the corresponding character. Assume the visual features $F_{g} \in R^{hw \times d_v}$, where $h$ and $w$ are the height and width of the feature map, and the semantic information from the self-attention is $E_{s_t}$. The cross-attention calculation is described as   \Cref{eq:ca} and   \Cref{eq:qkv}:

\begin{equation}
    E_{c_t} = Attention(Q_t, K, V) = Softmax(\frac{Q_tK^T}{\sqrt{d}})V,
    \label{eq:ca}
\end{equation}

\begin{equation}
    Q_t = W_QE_{s_t}, \quad K=W_KF_g, \quad V=W_VF_{g},
    \label{eq:qkv}
\end{equation}
where $W_Q \in R^{d\times d_e}$, $W_K \in R^{d\times d_v}$, $W_V \in R^{d\times d_v}$. The local features need to be adaptively extracted from $F_g$. Therefore, the grid window size is estimated by $E_{s_t}$ as   \Cref{eq:gs}:

\begin{equation}
    gs_t = \{(h_t, w_t)_l\} = Sigmoid(FC(E_{s_t})).
    \label{eq:gs}
\end{equation}

The grid size $gs_t \in R^{l\times 2}$ shows the height and width of the perceived range of local attention on different levels of feature maps. Suppose that the grid size is $g$. Then the coordinates of grid points can be formulated as  \Cref{eq:gp}:

\begin{equation}
    gp_{l_t} = \{(-\frac{w_{l_t}}{2} + \frac{w_{l_t}}{g-1}i, -\frac{h_{l_t}}{2} + \frac{h_{l_t}}{g-1}i) + p_t)\} \in R^{g\times g\times 2},
    \label{eq:gp}
\end{equation}
where $i, j=0,1,...,g-1$ represents the point in row $i$ and column $j$ of the grid. $p_t = (p_{t_x}, p_{t_y})$ is equivalent in different levels of feature maps due to the normalization of coordinates. The grid features can be obtained by bi-linear interpolation as   \Cref{eq:gf}:

\begin{equation}
F_{g_{t_l}} = Sample(F, gp_{l_t}) \in R^{g^2 \times d_v}.
    \label{eq:gf}
\end{equation}

Then visual features can be concentrated as  \Cref{concat}, where $n$ is the levels of feature maps and $n=4$.
\begin{equation}
F_{g} = Concat(F_{g_{t_0}}, F_{g_{t_1}}, ...,  F_{g_{t_n}}).
    \label{concat}
\end{equation}

In conclusion, the MAAM module calculates attention on different local text areas when decoding every text instance. This design not only ensures that the decoder perceives the visual information of characters but also prevents the high calculation consumption with the whole image area. The outputs of MAAM are fed into a regression branch to predict the coordinate shift $\Delta x_t$ and $\Delta y_t$ of the next character and a classification branch with a Softmax layer to predict the character. As the input of the next step, the coordinate shift and the content of the character in the current step appends into decoded character sequences. The auto-regressive process will be ended until the [EOS] token. Therefore, our spotter makes the decoder decide the ending position independently without sophisticated detection.

\subsection{Label Generation}
\label{label}
% 参考点标签实例
To ensure that the local grid features can be extracted from the corresponding characters in each decoding step, the prediction of reference points should be as accurate as possible, so it is necessary for reference points to be supervised directly. 
The ideal reference point should be the center of each character. However, due to the expensive cost of annotating character positions, most datasets do not contain character-level annotations, so it is not easy to obtain an accurate character-level center position. 
Fortunately, the local grid designed in MAAM can cover a certain image range and the grid size is adaptive and variable, so the local grid can adaptively perceive the corresponding feature information through the attention mechanism even if the character reference point's location could be inaccurate. Therefore, we propose a simple strategy of label generation using the existing annotations in the scene text dataset.

\begin{figure}[t]
\centering
\includegraphics[width=0.5\textwidth]{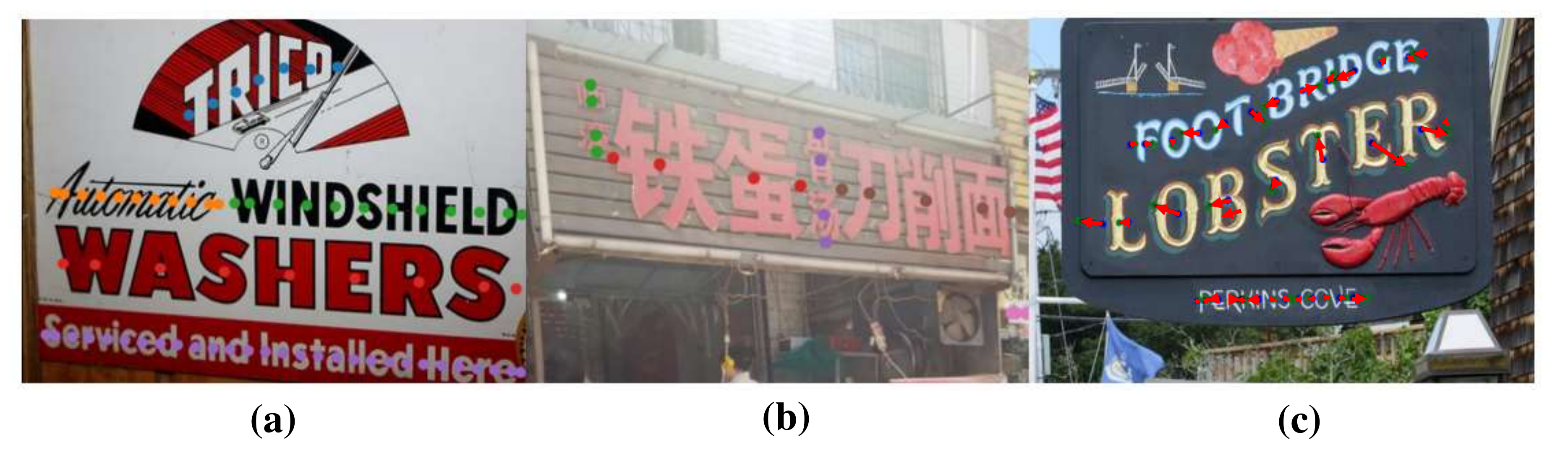}
    \caption{The visualization of Label Generation on different language datasets. Points in different colors represent the different text instances in (a) and (b). The red arrows in (c) show the disturbance shift of center points.}
    \label{fig:label}
\vspace{-15pt}
\end{figure}

Considering the existing datasets in scene text spotting use polygons as the representations of text boundaries, and the order of vertices follows the reading orders. Therefore, we obtain two edges for each text instance along the reading order. Given the length of the text instance $m$, two edges $E_{top} = \{t_i\}_{i=0}^m$ and $E_{bottom} = \{b_i\}_{i=0}^m$ are interpolated as $m+1$ points from raw polygons. The center line $C = \{c_i\}_{i=0}^m$ can be calculated as   \Cref{eq:centerline}:

\begin{equation}
    c_i = \frac{t_i + b_i}{2}.
    \label{eq:centerline}
\end{equation}

Suppose that each character is equal in width. The $t$-th character ($t=0,1,...,m-1$) center coordinate $r_t$ can be calculated as  \Cref{eq:centerxy}:

\begin{equation}
    r_t = \frac{c_t + c_{t+1}}{2}.
    \label{eq:centerxy}
\end{equation}

Now $r_t$ is regarded as the reference point of $t$-th character. During the auto-regressive decoding, $c_0$ and $c_m$ are the [SOS] and [EOS] tokens respectively for character position. The visualization of reference points is shown in \Cref{fig:label}.

During the training phase, the teacher-forcing training strategy in an auto-regressive manner alleviates the difficulty of optimization, which causes bad performances of our spotter. If the reference points $r_t$ are fed into the decoder during the training stage, while the reference points are predicted value $p_t$ during the inference stage, it will lead to exposure bias in the inference because of the accumulative errors of predicted reference points. Moreover, the precise annotations in the training stage make the model overlook the previous hidden states so that cannot learn the semantic knowledge. To solve these problems, we design a strategy for disturbing the reference points in the training phase. Given a set of reference points $R = \{r_t\}_{t=0}^n$, we use   \Cref{eq:disturb} to describe this procedure, where $\eta_x$, $\eta_y$ are the disturbing weight of x-axis and y-axis respectively, and distributed between -1 and 1 uniformly. This setting not only makes reference points have a certain disturbance but also prevents the reference point from disturbing too much. The visualization of disturbing reference points is shown in  \Cref{fig:label} (c). 

\begin{equation}
    r_t' = r_t + (\frac{\eta_x}{2} |r_{t_x} - r_{(t-1)_x}|, \frac{\eta_y}{2} |r_{t_y} - r_{(t-1)_y}|).
    \label{eq:disturb}
\end{equation}

\subsection{Optimization}

The overall loss function $\mathcal{L}$ of LSGSpotter includes two parts, the detection loss $\mathcal{L}_{det}$ and the decoder loss $\mathcal{L}_{dec}$. $\mathcal{L}$ can be represented as   \Cref{eq:loss_all}, where $\lambda$ is the weight factor for balancing between $\mathcal{L}_{det}$ and $\mathcal{L}_{dec}$:

\begin{equation}
    \mathcal{L} = \mathcal{L}_{det} + \lambda \mathcal{L}_{dec}.
    \label{eq:loss_all}
\end{equation}

% Except for $\mathcal{L}_{start}$ (different with E.2) defined as   \Cref{eq:start}, $\mathcal{L}_{det}$ also considers the Cross-Entropy Loss $\mathcal{L}_{text}$ for the classification of text areas, where $y_i$ and $\hat{y}_i$ are the round truth and prediction of text region. $\mathcal{L}_{text}$ and $\mathcal{L}_{det}$ are defined as   \Cref{eq:ltext} and   \Cref{eq:ldet} respectively.

% \begin{equation}
%     \mathcal{L}_{text} = -\sum_{i}y_i log \hat{y_i},
%     \label{eq:ltext}
% \end{equation}

% \begin{equation}
%     \mathcal{L}_{det} = \mathcal{L}_{start} + \mathcal{L}_{text}.
%     \label{eq:ldet}
% \end{equation}

In addition, $\mathcal{L}_{dec}$ consists of the reference regression loss and character recognition loss. It can be defined as   \Cref{eq:ldec}, where $y_t$ and $\hat{y}_t$ are ground truth and prediction of the transcription.

\begin{equation}
    L_{dec} = \sum_{t} [-y_t log \hat{y}_t + SmoothL1(\hat{p}_t, p_t)].
    \label{eq:ldec}
\end{equation}

\section{Experiments}

% \subsection{Datasets}

% \textbf{SynthText-150k}  \cite{liu2020abcnet} contains 150k images with texts for pre-training. It is generated by SynthText toolbox  \cite{gupta2016synthetic}, including curved texts and horizontal instances. 
    
% \noindent\textbf{ICDAR2015}  \cite{karatzas2015icdar} is an incidental scene text benchmark, which includes 1000 training images and 500 images for testing. It provides multi-oriented text instances with quadrangle annotations.

% \noindent\textbf{InverseText}  \cite{ye2023dptext} is a new benchmark to solve the reading order problem in the scene text spotting. It only consists of 500 testing images. It is a challenging arbitrarily shaped scene text test set with about 40\% inverse-like scene texts, and some of these texts are even mirrored.

% \noindent\textbf{ReCTS}  \cite{zhang2019icdar} includes 25k Chinese annotated signboard images, where 20k images are for the training set and the rest are for the testing set. In ReCTS, there are many more character classes, more complicated layouts, and various fonts than in English datasets. 

% \noindent\textbf{Total-Text}  \cite{ch2017total} includes arbitrary-shaped and focused text instances with word-level annotations. There are 1255 training images and 300 testing images.

% \noindent\textbf{SCUT-CTW1500}  \cite{yuliang2017detecting} includes arbitrary-shaped and focused text instances. Different from Total-Text, it is annotated with line-level. 

\begin{table*}[t]\setlength{\tabcolsep}{12pt}
\centering
\begin{tabular}{lcccccc}
\hline
\multirow{2}{*}{Methods} & \multicolumn{2}{c}{InverseText}                      & \multicolumn{2}{c}{SCUT-CTW1500}                         & \multicolumn{2}{c}{Total-Text}                     \\ \cline{2-7} 
                         & None                     & Full                     & None                     & Full                     & None                     & Full                     \\ \hline
\multicolumn{7}{c}{Box/Polygon-based Metric}                                                                                                                                                       \\ \hline
TextDragon  \cite{feng2019textdragon}            &      -                        & -                & 39.7                     & 72.4                     &   44.8                     & 74.8                      \\
% CharNet   \cite{xing2019convolutional}                & 66.6                     & -                        & -                        & -                        & -                        & -                        \\
ABCNet   \cite{liu2020abcnet}                 & 22.2                     & 34.3                      & 45.2                     & 74.1                     &   64.2                     & 75.7                  \\

TextPerceptron  \cite{qiao2020text}          &     -                        & -                 & 57.0                       & -                        & 69.7                     & 78.3                        \\
% PGNet   \cite{wang2021pgnet}                  & 60.5                     & -                        & -                        & -                        & -                        & -                        \\
MaskTextSpotterV3  \cite{liao2020mask}       &     -                        & -                  & -                        & -                        & 71.2                     & 78.4                       \\
MANGO   \cite{qiao2021mango}                  & -                        & -                     & 58.9                     & 78.7                     &  72.9                     & 83.6                       \\
PAN++   \cite{wang2021pan++}                  &   -                        & -                  & -                        & -                        & 68.6                     & 78.6                         \\
ABCNetv2  \cite{liu2021abcnet}                &    34.5                     & 47.4                   & 57.2                     & 77.2                     & 70.4                     & 78.1                    \\

Boundary  \cite{lu2022boundary}                &    -                        & -                   & -                        & -                        & 65.0                       & 76.1                       \\
TPSNet  \cite{wang2022tpsnet}                  &   -                        & -                    & 60.5                     & 80.1                     & 78.5                     & 84.1                       \\

SwinTextSpotter  \cite{huang2022swintextspotter}         &        55.4                     & 67.9              & 51.8                     & 77                       & 74.3                     & 84.1                     \\
TESTR  \cite{zhang2022text}                   &    34.2                     & 41.6                  & 56.0                       & 81.5                     & 73.3                     & 83.9                     \\
UNITS  \cite{kil2023towards}                   & -                        & -                       & \underline{66.4}                     & 82.3                     &   78.7                     & 86.0                      \\
DeepSOLO  \cite{ye2023deepsolo}                &   64.6                     & 71.2                      & 64.2                     & 81.4                     & 79.7                     & 87.0                    \\
ESTextSpotter  \cite{huang2023estextspotter}           &  -                        & -                     & 64.9                     & 83.9                     & \underline{80.8}                     & \underline{87.1}                       \\
IAST  \cite{zhang2024inverse}                    & \underline{68.8} & \underline{80.6}& {62.4} & {82.9} & {71.9} & {83.5}  \\ \hline
\multicolumn{7}{c}{Point-based Metric}                                                                                                                                                     \\ \hline
SPTS   \cite{peng2022spts}                   & 38.3                     & 46.2                        & 63.6                     & 83.8                     &    74.2                     & -                  \\
SPTS v2   \cite{liu2023spts}                & {63.4} & {74.9} & {63.6} & \underline{84.3} & {75.5} & {84.0} \\
LSGSpotter               &   \textbf{73.7}                     & \textbf{82.3}                   &     \textbf{68.9}                     &        \textbf{84.4}                  &   \textbf{81.5}                     & \textbf{87.3}                   \\ \hline
\end{tabular}
\caption{End-to-end scene text spotting results on the InverseText, Total-Text, SCUT-CTW1500 English benchmarks. Bold indicates SOTA, and underline indicates the second best. “None” represents lexicon-free, while “Full” indicates all the words in the test set are used.}
\label{tab:sota}
\vspace{-15pt}
\end{table*}

\subsection{Settings}
Following the settings of previous works, we pre-train our model on SynthText-150k, MLT-2017 \cite{nayef2017icdar2017}, ICDAR2013 \cite{karatzas2013icdar}, ICDAR2015 \cite{karatzas2015icdar}, TextOCR \cite{singh2021textocr} and Total-Text for 600k iterations, which AdamW optimizes with the learning rate of 2e-4 and the weight decay is 1e-4. After pretraining, the model is fine-tuned on the training split of the target benchmark for 200 epochs. The initial learning rate is 1e-4 and declined to 1e-5 on the 60th epoch. The entire model is trained on 4 NVIDIA RTX3090 GPUs with a batch size of 4 on the single GPU. In addition, we utilize the ResNet50 \cite{he2016deep} with deformable convolution module \cite{dai2017deformable} for the backbone and the 6-layer Transformer decoder for the auto-regressive stage. During the training, the short size of an input image is resized and padded to 960. Random cropping and rotating are employed for data augmentation. In the inference stage, we resize the short edge to 960 while keeping the long side shorter than 1600 with the fixed aspect ratio. 
For evaluation, we follow the point-based evaluation metrics proposed by SPTS \cite{peng2022spts}, because our method only generates center points for representing the position information rather than accurate polygons. SPTSv2 \cite{liu2023spts} has proven the fairness of point-based metrics compared with the box-based.

\begin{figure}[t]
\centering
\includegraphics[width=0.48\textwidth]{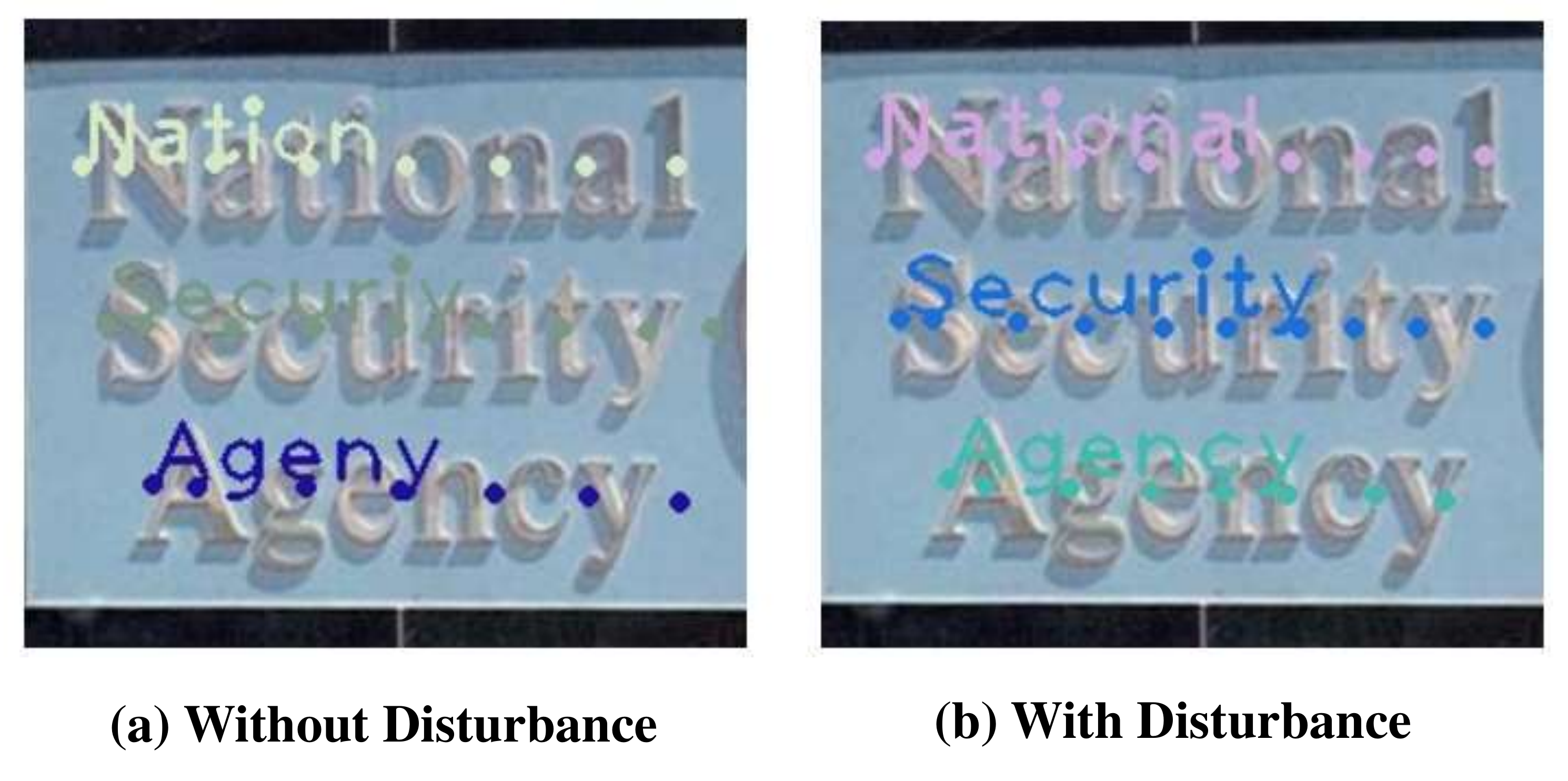}
    \caption{Impact of reference point disturbance strategy on model performance. Without disturbance during training, some characters will be omitted in the inference stage.}
    \label{fig:disturb}
    \vspace{-10pt}
\end{figure}

\subsection{Comparison with State-of-the-art Methods}
Illustrated by \Cref{tab:sota}, our method achieves state-of-the-art results on the InverseText, the most challenging benchmark. Specifically, it presents 73.7\% performance without the help of lexicons, being 4.9\% better than IAST \cite{zhang2024inverse}, a scene text spotter specifically designed for inverse texts. Our spotter also achieves 82.3\% performance on evaluation in ``Full" condition. The main reason is that LSGSpotter has SPLM to locate the start point. It not only learns where the text instance is located but is also implicitly aware of the reading order. The local semantics guidance also leverages fine-grained information to determine the right reading order.

Furthermore, We report experiment results on several public benchmarks, including English benchmarks InverseText, Total-Text, and SCUT-CTW1500. As shown in  \Cref{tab:sota}, our spotter outperforms with 81.5\% on Total-Text and 68.9\% on SCUT-CTW1500, 0.7\% higher than ESTextSpotter \cite{huang2023estextspotter} and 2.5\% higher than UNITS \cite{kil2023towards}. With the help of the lexicon, our spotter also significantly surpasses the previous methods on the ``Full" evaluation protocol. 
% We believe that the fine-grained semantic information has been learned in an auto-regressive manner. 
We believe that the auto-regressive manner aids in learning effective and fine-grained semantic information to decode the scene texts accurately.
% \subsubsection{Comparison on Chinese benchmarks}

% To evaluate the performance of Chinese scene text spotting, we make a few adjustments to obtain the whole text boundaries rather than the centerline. Based on predicting the shift of the next character,  we add the shift prediction related to the upper edge of each character in the decoder. Then, we can get all points of text boundaries and external polygons of all text instances. As shown in  \Cref{table:Chinese}, our end-to-end spotting result surpasses the previous spotters. It also proves that our spotter is effective for Chinese scene text spotting.

% \subsubsection{Comparison on English benchmarks}

% Moreover, LSGSpotter performs the competitive results on the curve datasets Total-Text and SCUT-CTW1500 as shown in  \Cref{table:english}. We notice our method performs slightly worse on English datasets than ESTextSpotter. We argue that the heuristics and segmentation-based algorithm of start point localization in SPLM could lack sensitivity for small text instances. The simple fusion manner in SPLM is also the main reason. 
% In addition, insufficient the setting of iterations during pretraining is another reason.

\begin{table}[t]
\begin{tabular}{ccc}
\hline
Settings & Total-Text & SCUT-CTW1500 \\ \hline
Without disturbance                   & 77.8       & 62.6          \\
   With disturbance                   & 81.5 (+3.7)      & 68.9 (+6.3)       \\ \hline
\end{tabular}

\caption{ Ablation experiments about reference point disturbance}
\label{table:pp}
\vspace{-15pt}
\end{table}

% \usepackage{multirow}
\begin{table}[t]

\setlength{\tabcolsep}{18pt}
\centering
\begin{tabular}{ccccc}
\hline
$F_{r}$ & $F_c$ & $F_s$ & Total-Text  \\
\hline
           \checkmark                   &                  &                    &        76.4                                        \\
              \checkmark                &     \checkmark               &                  &          81.1  (+4.7)                                          \\
               \checkmark               &                  &       \checkmark             &                80.4 (+4.0)                                              \\ 
                              \checkmark              &           \checkmark        &       \checkmark             &    81.5  (+5.1)                                                    \\ 
               
               \hline
\end{tabular}
\caption{ Ablation experiments about the approach of starting point localization. $F_r$, $F_c$, $F_s$ are the text region map, the text centerline map, and the start point map described in  \Cref{SPLM} in detail. The evaluation protocol is ``None".}
\label{table:splm}
\vspace{-5pt}
\end{table}

% \usepackage{multirow}
\begin{table}[t]
\centering

\begin{tabular}{ccccccc}
\hline
\multirow{2}{*}{Grid} & \multirow{2}{*}{AS} & \multicolumn{3}{c}{Grid Size} & \multirow{2}{*}{CTW1500} & \multirow{2}{*}{FPS} \\ \cline{3-5}
                      &                                 & 3×3      & 5×5      & 7×7     &                          &                      \\ \hline
                      &                                 &          &          &         & 58.6                     & 1.1                  \\
\checkmark                      &                                 &          &    \checkmark        &         & 63.7 (+5.1)                    & 7.6                  \\
         \checkmark               &         \checkmark                          &          &    \checkmark        &         & 68.9     (+10.3)                & 7.4                  \\
           \checkmark             &       \checkmark                            &   \checkmark         &          &         & 67.5 (+8.9)                   & 8.4                  \\
             \checkmark            &           \checkmark                         &          &          &   \checkmark         & 69.2 (+10.6)                  & 6.9                  \\ \hline
\end{tabular}

\caption{ Ablation experiments about MAAM. AS means Adaptive Scale. The evaluation protocol is ``None".}
\vspace{-15pt}
\label{table:maam}
\end{table}

\begin{table}[t]
            % \vspace{-10px}
             \centering
            \begin{tabular}{cc}
            \hline
             \multicolumn{1}{l}{Settings}                      & \multicolumn{1}{c}{FLOPs} \\ \hline
            \multicolumn{1}{l}{LSGSpotter}                   & 194.47G                   \\
            - Grid Sampling & 846.35G                   \\ \hline
            \end{tabular}
            \caption{Ablations for computational consumption. ``LSGSpotter" indicates the default setting and "-Grid Sampling" refers to the configuration of LSGSpotter without grid sampling. }
\label{tab:computational}
% \vspace{-15pt}
\end{table}

\begin{table}[t]
\centering
\setlength{\tabcolsep}{3pt}
\begin{tabular}{
lcccccccc}
\hline
{$T$} &
  { 0.2} &
  { 0.3} &
  { 0.4} &
  { 0.5} &
  { 0.6} &
  { 0.7} &
  { 0.8} &
  { 0.9} \\
  \hline
% {P} &
%   { 78.4} &
%   { 83.8} &
%   { 87.5} &
%   { 89.7} &
%   { 91.0} &
%   { 93.5} &
%   { 95.1} &
%   { 96.2} \\
% {R} &
%   { 99.1} &
%   { 98.1} &
%   { 87.0} &
%   { 96.5} &
%   { 94.2} &
%   { 90.8} &
%   { 92.6} &
%   { 60.8} \\
{F} &
  { 87.5} &
  { 90.4} &
  { 92.0} &
  { 93.0} &
  { 92.6} &
  { 92.1} &
  { 88.4} &
  { 74.5} \\
  % \hline
{None} &
  { 80.0} &
  { 81.0} &
  { 81.3} &
  { 81.5} &
  { 81.4} &
  { 81.1} &
  { 79.1} &
  { 68.9} \\
  \hline
\end{tabular}
\caption{Ablation for the threshold $T$ of SPLM. ``P", ``R", ``F" and ``None" denote precision, recall, and F1-score for detection and end-to-end spotting performance, respectively.}
\label{tab:T}
\vspace{-15pt}
\end{table}

\subsection{Ablations}
\subsubsection{The disturbance of reference points}

 \Cref{table:pp} shows the significant influence of the disturbance of reference points. "With disturbance" expresses that the reference points are disturbed during training. The experiment results show the disturbance of reference points surpasses the ``Without disturbance" by 3.7\% on Total-Text and 6.3\% on SCUT-CTW1500. We explain this improvement by  \Cref{fig:disturb}. Whether the reference disturbance is used or not, the prediction of coordinates will inevitably have slight errors. The error accumulations in an auto-regressive manner could lead to missing some characters, as shown in \Cref{fig:disturb}(a). However, when we use the strategy of reference point disturbance, the model can learn some linguistic knowledge to alleviate the influence of reference shift. Therefore, the design of reference points disturbance during training is necessary.

\subsubsection{Ablation on the settings of SPLM}

In the SPLM, we use three feature maps to locate the reference position. In the ablation study on SPLM, we analyze the necessity of three different features. The experiment results are shown in  \Cref{table:splm}. When we only use the $F_r$ to locate the reference point, the F-measure is 76.4\%  on Total-Text. When we leverage the $F_c$ with $F_r$, it improves by 4.7\% compared with the baseline. After replacing $F_s$ with $F_c$, it further enhances the performance of 4.0\%. Eventually, we use three feature maps together and obtain 5.1\% better than the baseline. The above experiment results prove the effectiveness of the fusion of three feature maps at the same time.

\subsubsection{Ablation on the settings of MAAM}

 \Cref{table:maam} shows the ablation studies results on MAAM, in which ``Grid" indicates the utilization of the local grid, and ``AS", the abbreviation of ``Adaptive Scale", expresses whether the grid scale is learnable adaptively. In the event of dropping out of the local grid, it is notable that all image tokens originating from multi-level feature maps will be engaged in the computation of cross-attention mechanisms throughout the decoding phase corresponding to each character. As shown in the 1st and 2nd lines of  \Cref{table:maam}, dense attention cannot bring satisfying performance and efficient inference speed. The comparison of the 2nd and 3rd lines of  \Cref{table:maam} validates the performance and efficiency of the adaptive scale. We explain that the adaptive scale makes the decoder perceive the location of each character due to the large-scale variation range of scene texts. Additionally, we quantitatively compare the complexity reduction caused by grid sampling in \Cref{tab:computational}. The sampling operation brings about several times increase in efficiency.   Furthermore, we attempt three different grid number settings to explore the effect of grid size. The experiments claim that the grid size of $7 \times 7$ brings the most enhancement on SCUT-CTW1500. However, it is only a 0.3\% improvement compared with the settings of $5 \times 5$, but it slows down the inference speed extremely. Therefore, we use $5 \times 5$ as the default setting of the grid size. 

\subsubsection{Ablation on the settings of \textit{T}}
As shown in \Cref{tab:T}, we perform an ablation experiment on Total-Text using different values of $T$. Results show that when $0.3 \leq T \leq 0.7$, $T$ slightly impacts detection and spotting performances, because it serves as a threshold for selecting starting points derived from the center of connected regions, which offers robustness to noise. However, when $T \ge 0.8$, recall significantly drops, adversely affecting end-to-end results. Conversely, $T \leq 0.3$, false positives increase significantly, but end-to-end performance remains stable. This suggests that MAAM effectively filters out false positives by directly outputting the [EOS] token upon encountering such points, demonstrating its noise resistance.

\begin{figure}[t]
\centering
\includegraphics[width=0.45\textwidth]{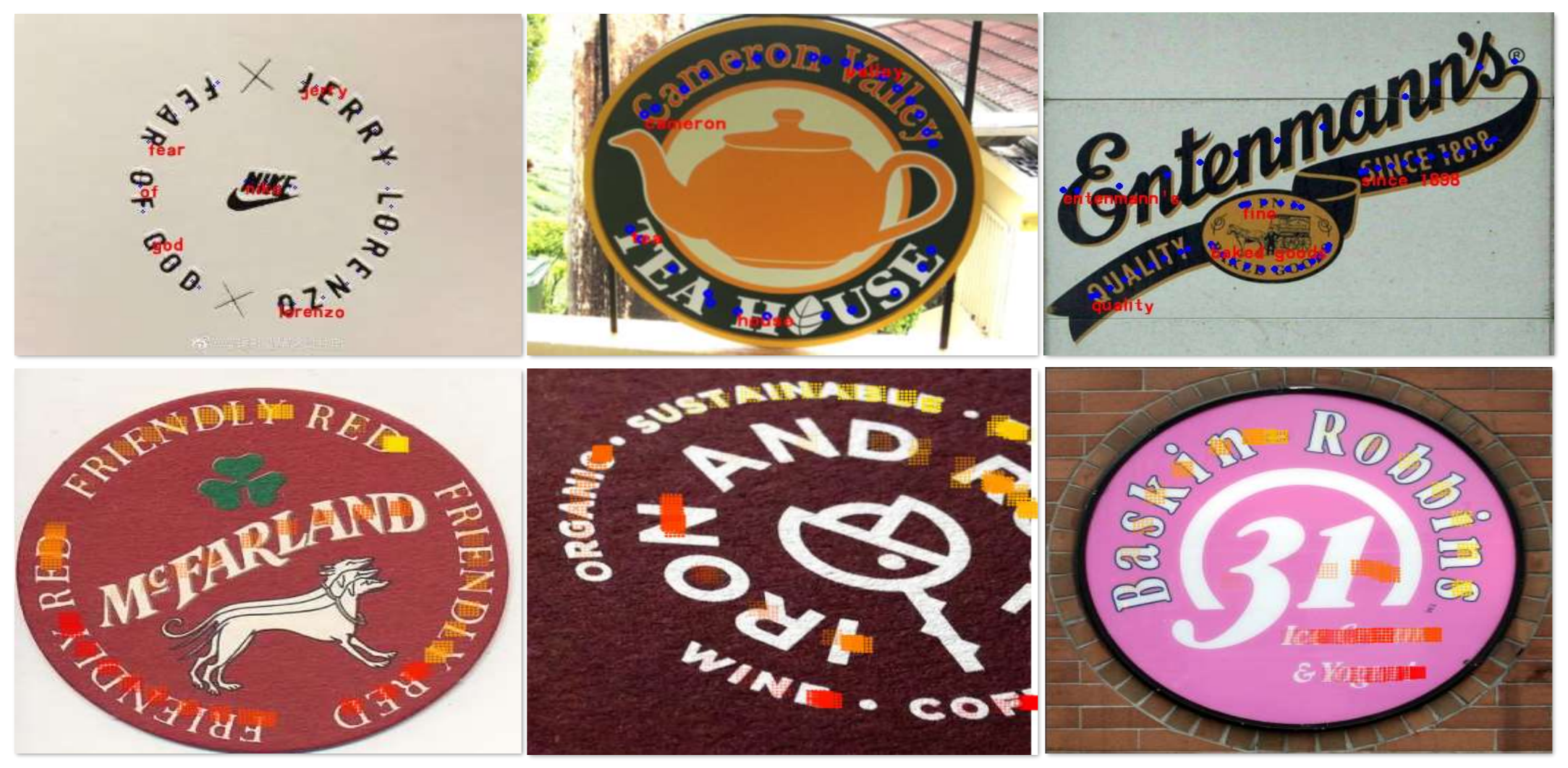}
    \caption{Qualitative results on the testing set of InverseText, Total-Text, SCUT-CTW1500 from left to right in the first line. The second line is the visualization of local grids predicted in MAAM for some challenging text instances. The color from light to deep indicates the decoding order.}
    \label{fig:vis}
\vspace{-15pt}
\end{figure}

\subsection{Visualization and Qualitative Analysis}

 \Cref{fig:vis} shows the visualization of four public benchmarks mentioned in this paper. It can
be observed that our method can accurately locate the start point and recognize the texts. Notably, our method performs well in large aspect-ratio, inverse, and curved text instances by locating the start point and predicting the next character in an auto-regressive manner, which helps determine the reading order of text instances.

\section{Conclusion}
In this paper, we propose LSGSpotter, a local-semantics-guided scene text spotter. To address the extreme dependency of accurate detection in the whole spotting pipeline, we revisit the recognition process and propose two effective modules, SPLM and MAAM respectively. SPLM locates the start point of the text instance to avoid our spotter paying more attention to the background noise. Moreover, SPLM learns implicit reading orders of text instances and solves the inverse text effectively. MAAM decodes the character shift and content step-by-step. The adaptive local grid attention can save the computational resources further. Extensive experiments show the outperformed performances of our spotter on three English benchmarks. It proves our spotter can handle the arbitrary reading order problem effectively. 
% However, there are two main limitations of our method. First, our approach leverages an auto-regressive manner to emphasize semantic context but struggles with contextless words. Second, LSGSpotter fails to detect mirror-inverted text, which we attribute to the limited occurrence of this pattern in training datasets. 
In the future, We hope our method can inspire further exploration of detection-free spotting.

\section{Acknowledgments}

Supported by the National Natural Science Foundation of China (Grant NO 62376266 and 62406318), and by the Key Research Program of Frontier Sciences, CAS (Grant NO ZDBS-LY-7024).

\bibliography{aaai25}

\newpage

\appendix

\section{Datasets}

\textbf{SynthText-150k}  \cite{liu2020abcnet} contains 150k images with texts for pre-training. It is generated by SynthText toolbox  \cite{gupta2016synthetic}, including curved texts and horizontal instances. 
    
% \noindent\textbf{ICDAR2015}  \cite{karatzas2015icdar} is an incidental scene text benchmark, which includes 1000 training images and 500 images for testing. It provides multi-oriented text instances with quadrangle annotations.

\noindent\textbf{InverseText}  \cite{ye2023dptext} is a new benchmark to solve the reading order problem in the scene text spotting. It only consists of 500 testing images. It is a challenging arbitrarily shaped scene text test set with about 40\% inverse-like scene texts, and some of these texts are even mirrored.

% \noindent\textbf{ReCTS}  \cite{zhang2019icdar} includes 25k Chinese annotated signboard images, where 20k images are for the training set and the rest are for the testing set. In ReCTS, there are many more character classes, more complicated layouts, and various fonts than in English datasets. 

\noindent\textbf{Total-Text}  \cite{ch2017total} includes arbitrary-shaped and focused text instances with word-level annotations. There are 1255 training images and 300 testing images.

\noindent\textbf{SCUT-CTW1500}  \cite{yuliang2017detecting} includes arbitrary-shaped and focused text instances. Different from Total-Text, it is annotated with line-level.

\section{The Visualization of label generation}
In the Start Point Localization Module (SPLM), we leverage three kinds of ground truth,  text region $GT_r$, text center $GT_c$, and start points $GT_s$, which are intuitively shown in the \Cref{fig:label}. It is hard for text regions to separate texts but text centers conduct it easily. Start points are the first reference point, implicitly indicating the reading order. In addition, reference points in (d) of \Cref{fig:label} are used to calculate the loss of the decoder, which is referred to as Equation (18).

\begin{figure}[h]
\centering
\includegraphics[width=0.5\textwidth]{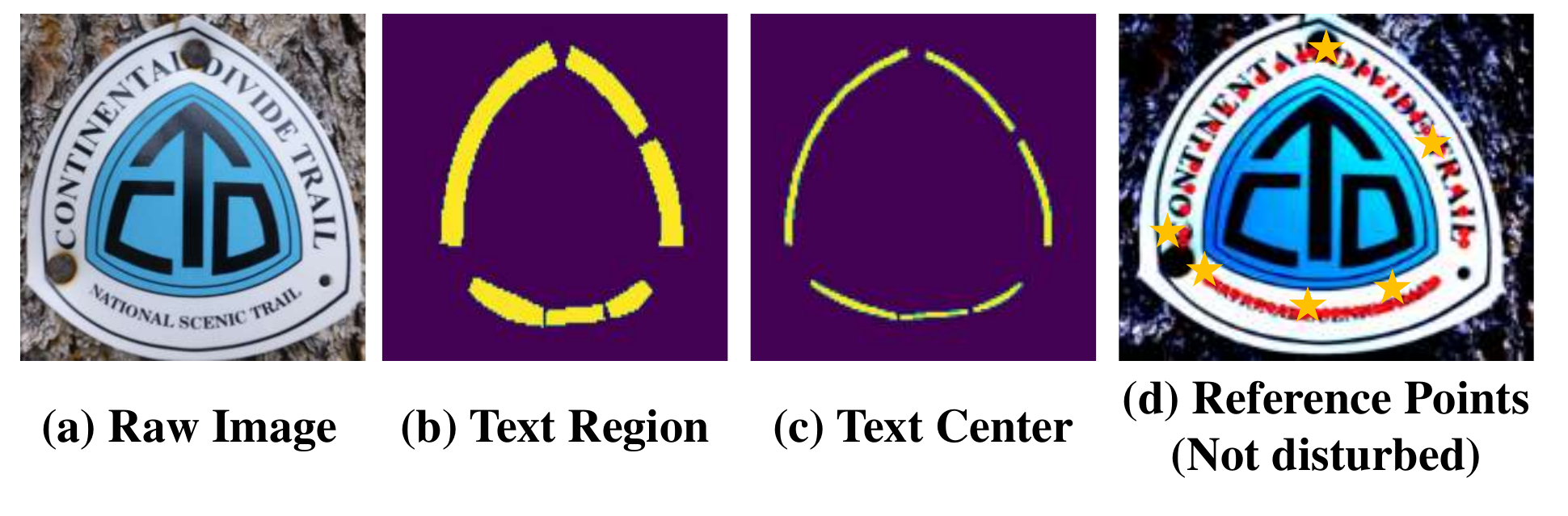}
    \caption{The visualization of ground truths for text region $GT_r$, text center $GT_c$, start points $GT_s$ and reference points. Specifically, Orange stars in (d) are start points, and red points in (d) are reference points that are not disturbed.}
    \label{fig:label}
\end{figure}

\section{More ablation studies}

\subsection{The Ablation for $\lambda$}
The weight factor $\lambda$ aims to balance the optimization between SPLM and MAAM. Here we conduct experiments for different settings of $\lambda$ to analyze the impact of $\lambda$, as shown in \Cref{table:lambda}. The experiment results indicate the best performance can be achieved when setting $\lambda=1$. Therefore, the weight factor $\lambda$ is set to 1 by default if no special statement exists.

\subsection{The Ablation for fine-grained feature}
We introduce DEER\cite{kim2022deer}, a scene text spotter with single-point instance localization, to compare the performance between fine-grained and single-point localization. The results on Total-Text are shown as \Cref{tab:3}, which indicates the significant increase in replacing the single-point with fine-grained localization supervision.

\begin{table}[t]\setlength{\tabcolsep}{16pt}
\caption{The ablation study for the setting of $\lambda$. Bold indicates the best performance. ``None” represents lexicon-free, while “Full” indicates all the words in the test set are used.}
\begin{tabular}{ccccc}
\hline
\multirow{2}{*}{$\lambda$} & \multicolumn{2}{c}{Total-Text} & \multicolumn{2}{c}{Inverse-Text} \\ \cline{2-5} 
    & None           & Full          & None            & Full           \\ \hline
0.1 &       80.1         &       86.6        &        72.9         &      81.3            \\
0.5 &      \textbf{81.7}          &     87.1          &      73.4           &      81.4          \\
1   &      81.5         &     \textbf{87.3}          &       \textbf{73.7}          &     \textbf{82.3}           \\
5   &      80.9          &     87.0          &          72.9       &      81.6          \\
10  &      80.3          &     86.8          &       72.4          &    80.7          \\ \hline
\end{tabular}
\label{table:lambda}
\end{table}

\begin{table}[t]
\caption{Ablations for localization representations}
\centering
 % \vspace{-10px}
\begin{tabular}{lcc}
\hline
\multirow{2}{*}{Methods} & \multicolumn{2}{c}{Total-Text}   \\ \cline{2-3} 
                         & \multicolumn{1}{c}{None} & Full \\ \hline
DEER                     & 74.8                      & 81.3 \\
LSGSpotter (ours)        & 81.5 (+6.7)                     & 87.3 (+6.0) \\ \hline
\end{tabular}
\label{tab:3}
% \vspace{-15pt}
\end{table}

\section{The upper bound}
Considering that the SPLM could omit some words, leading to failure detection, we explore the upper bound of our method. Specifically, we replace the start point predicted by SPLM with one generated by ground truth. This operation eliminates the effect of a low recall rate from SPLM. \Cref{table:upper} shows the upper bound of our spotter achieves performance of 84.0$\%$ on Total-Text. This result claims that our spotter has great potential for general scene text spotting.

\begin{table}[h]
\centering
\caption{The upper bound of LSGSpotter. \textit{Pred start point} indicates the start points are predicted by SPLM. \textit{GT start point} represents that the start points are generated from ground truth.}
\begin{tabular}{lcc}
\hline
\multirow{2}{*}{Setting} & \multicolumn{2}{c}{Total-Text}   \\ \cline{2-3} 
                         & \multicolumn{1}{c}{None} & Full \\ \hline
Pred start point         & 81.5                      & 87.3 \\
GT start point           & 84.0                      & 89.6 \\ \hline
\end{tabular}
\label{table:upper}
\end{table}

\begin{figure}[t]
\centering
\includegraphics[width=0.48\textwidth]{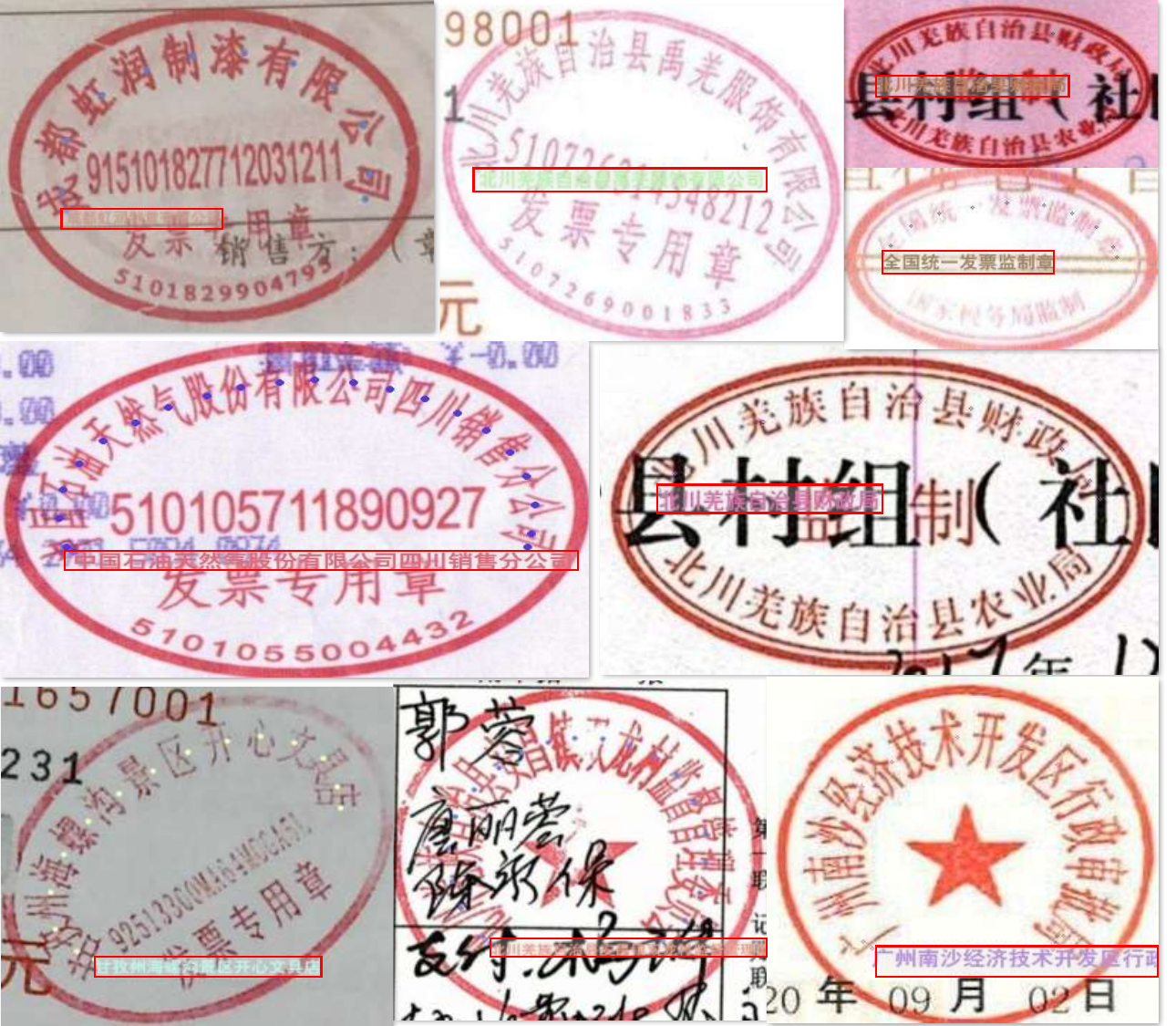}
    \caption{The visualization on ICDAR2023-ReST. The red boxes aim to emphasize the predicted transcription.}
    \label{fig:rest}
\end{figure}

\section{Visualization}

To validate the robustness
in practical applications, we test our model on ICDAR2023-ReST \cite{yu2023icdar},  a challenging dataset suffering background noises and overlapped texts. ICDAR2023-ReST aims to extract the title of the seal. \Cref{fig:rest} shows the qualitative results. Qualitative results indicate our spotter has prominent noise resistance for overlapped and curved texts.

\section{Discussion of Limitations}
There are two main limitations of our method. First, our approach leverages an auto-regressive manner to emphasize semantic context but struggles with contextless words. Second, LSGSpotter fails to detect mirror-inverted text, which we attribute to the limited occurrence of this pattern in training datasets. 

\end{document}